\documentclass[english]{lni}

\usepackage{graphicx}
\usepackage{fancyhdr}
\usepackage{changepage} %for changing topmargin on first page
\usepackage{listings} %if lstlistings is used
\usepackage{booktabs}
\usepackage{tikz}
\usepackage{pgfplots}
\usepackage{tabularray}
\pgfplotsset{compat=1.17}
\usepackage[figurename=Fig., tablename=Tab., small]{caption}

\fancypagestyle{titlepage}{
\fancyhead[RO]{} % do NOT modify these lines
\fancyfoot{}}

\renewcommand{\headrulewidth}{0.4pt} %horizontal line below header
\author{Luiz F. P. Southier\footnote{Federal University of Technology - Parana (UTFPR), Brazil, luizsouthier@utfpr.edu.br}, 
Marcelo Filipak,
Luiz A. Zanlorensi,
Ildefonso Wasilevski,
Fabio Favarim,
Jefferson T. Oliva,
Marcelo Teixeira,
Dalcimar Casanova }

\title{An on-production high-resolution longitudinal neonatal fingerprint database in Brazil}
\begin{document}

\maketitle

\renewcommand{\refname}{References}
\setcounter{footnote}{2} %Change to the number of authors for a correct numbering of the foot notes
\thispagestyle{titlepage}
%header setting after the second page
\pagestyle{fancy}
\fancyhead{} % clears header settings
\fancyhead[RO]{\small A longitudinal neonatal fingerprint database\hspace{25pt}  \hspace{0.05cm}}
\fancyhead[LE]{\hspace{0.05cm}\small  \hspace{25pt} Luiz F. P. Southier et. al.}
\fancyfoot{} % clears all footer settings
\renewcommand{\headrulewidth}{0.4pt} %line below header

\begin{abstract}
The neonatal period is critical for survival, requiring accurate and early identification to enable timely interventions such as vaccinations, HIV treatment, and nutrition programs. Biometric solutions offer potential for child protection by helping to prevent baby swaps, locate missing children, and support national identity systems. However, developing effective biometric identification systems for newborns remains a major challenge due to the physiological variability caused by finger growth, weight changes, and skin texture alterations during early development. Current literature has attempted to address these issues by applying scaling factors to emulate growth-induced distortions in minutiae maps, but such approaches fail to capture the complex and non-linear growth patterns of infants.
A key barrier to progress in this domain is the lack of comprehensive, longitudinal biometric datasets capturing the evolution of neonatal fingerprints over time. This study addresses this gap by focusing on designing and developing a high-quality biometric database of neonatal fingerprints, acquired at multiple early life stages. The dataset is intended to support the training and evaluation of machine learning models aimed at emulating the effects of growth on biometric features. We hypothesize that such a dataset will enable the development of more robust and accurate Deep Learning-based models, capable of predicting changes in the minutiae map with higher fidelity than conventional scaling-based methods. Ultimately, this effort lays the groundwork for more reliable biometric identification systems tailored to the unique developmental trajectory of newborns.
\end{abstract}
\section{Introduction}\label{sec:intro}

% CONTEXTO
Every year, more than 134 million children are born worldwide, resulting in more than 656 million children under the age of five in 2022 \cite{population}. Despite commendable efforts and advancements over the past decades, the imperative to enhance child survival persists. The year 2021 alone witnessed approximately 13,800 deaths per day of children under the age of five, an alarming statistic underscoring the pressing need for improved child care and survival strategies \cite{ufivemortality}.% Particularly vulnerable during the neonatal period, the first 28 days of life pose a critical window for a child's survival.

The lack of efficient identification, as highlighted by the World Health Organization (WHO), pose challenges in accurately tracking vaccination schedules \cite{ia2030}. The consequences are evident in the stagnation of global vaccination coverage, with 14.3 million children not receiving vaccinations in 2022, compared to 12.9 million in the pre-pandemic era \cite{immunizationcoverage}. Furthermore, 1.5 million children succumb to vaccine-preventable diseases each year \cite{reventabledeaths}. In regions with a high prevalence of HIV, timely diagnosis and intervention for HIV-exposed infants are crucial, and biometric solutions can aid in identifying and tracking these infants, ensuring prompt access to life-saving treatments \cite{sirirungsi2016early}.
In nutrition programs, the importance of neonatal identification is underscored by reports of fraud diverting food meant for needy children \cite{yemen}. 
Biometric data could offer a solution by facilitating the monitoring of nutrition programs, allowing for early interventions, and promoting healthy growth by tracking weight gain and vital metrics in children \cite{grantham2014effects}.
From a child protection standpoint, an efficient infant identification system would mitigate baby swaps in hospitals, locate missing or abducted children, and enhance airport security to prevent child trafficking \cite{sedlak2002national}. Neonatal biometrics also play a crucial role in national identity programs, ensuring secure and lifelong identities for children.

% PROBLEMA
Despite the consensus on the need for neonatal identification systems, neonates' delicate skin and tiny fingers pose challenges for conventional biometric systems designed for adult use \cite{jain2014recognizing}.
Adding to the complexity, in the initial five days post-birth, individuals undergo a daily body weight loss of 1-2\%, accompanied by a reduction in extracellular water from 40\% to 30\% \cite{lindower2017water,sharma2011adaptation}. Preterm infants may experience even more pronounced effects \cite{visscher2015newborn}, leading to a profound shift in body composition that influences the composition and thickness of the fingers skin layers \cite{visscher2015newborn,finlayson2022depth}.
Additionally, the rapid changes and growth in the children's fingers in the early stages of life impose an extra challenge referred to by the literature as the \textit{ageing effect} \cite{galbally2018fingerprint}. The ageing effect refers to the fingerprint quality and recognition performance that varies as the time lapse between the reference fingerprint sample, usually acquired at early hours of life, and the probe fingerprint sample, collected at older ages, changes. This effect is primarily related to the concept of fingerprint permanence, which refers to the ability of fingerprints to remain consistent and recognizable over time \cite{galbally2018study}.

The biometric systems use a set of key points in the fingerprint, referred to as \textit{minuatiae map},  to perform the recognition. To obtain a minutiae map from a fingerprint, first, the captured fingerprint needs to be segmented. The segmented image differentiates the raised lines of the finger's skin, named \textit{ridges}, from the depressed or low areas between the ridges, named \textit{valleys}. 
The minutiae map is defined by the points where ridges or valleys terminate.
%Each minutia $m_i$ in the minutiae map $M=\{m_0,...,m_i\}$ is represented by a tuple in the form of $m_i=(x_i,y_i,\theta_i)$, being $x_i$ and $y_i$, and $\theta_i$ the cartesian coordinates and the angle for the i-th minutia, respectively.
The changes due to the finger's growth, body weight loss, and skin composition alterations cause the minutiae map from a fingerprint to distort as time passes. These distortions, while minor for short time lapses become increasingly detrimental to the biometric system's recognition rate for more extended time lapses \cite{jain2014recognizing,camacho2017recognizing,jain2016fingerprint,galbally2018study,engelsma2021infant}.
% solution
To overcome this challenge, the biometric systems would benefit from a computational solution that could emulate the minutiae map from a fingerprint after some timelapse.
%Specifically, given a reference minutiae map $M_y$ collected at an age $y$ and a time-lapse $\Delta t$, the solution $f$ would emulate the fingerprint ageing and calculate the aged minutia map $M_{y'}$ at an age $y'$, such that $y' = y + \Delta t$, and that $M_{y'} = f(M_y)$.

% O QUE A LITERATURA JÁ TENTOU
Recognizing this gap, the literature has presented approaches to emulate the changes in the minutiae map by applying scaling factors. 
Some studies apply a fixed scaling factor \cite{jain2016fingerprint}, which proves suboptimal compared to using a scaling factor based on age category \cite{camacho2017recognizing}. While two-factor growth models have been suggested \cite{haraksim2019fingerprint,markert2019detecting}, they have not been evaluated on younger individuals. 

% GAP
While applying scaling factors has mitigated the ageing effect in the reported studies, this strategy seems to be an oversimplification because it assumes that fingerprints have a linear growth that such factors can model. However, as reported in \cite{schneider2010quantifying}, young children tend to exhibit distinctive growth patterns distorting the minutiae map in other ways. Additionally, these studies only use the minutiae maps as input for performing the ageing emulation. 
However, since young children tend to have a low-quality minutiae map
\cite{camacho2017recognizing}, the quantity and quality of minutiae are largely affected, and additional information, such as the ridge and valley configuration that defines the minutiae relation, would hypothetically benefit the emulation solution.

% PROBLEM AND OBJECTIVE
However, one of the fundamental barriers to advancing neonatal biometric research and developing robust identification systems is the lack of longitudinal biometric datasets—that is, datasets capturing biometric samples from the same individuals over extended periods of time. The absence of such data severely limits the ability to systematically study the effects of neonatal growth, morphological changes, and the aging effect on biometric recognition performance. As a result, existing studies are often constrained to cross-sectional or short-term data, which cannot accurately capture the temporal dynamics critical to fingerprint permanence and biometric reliability in infants.

To address this gap, the primary objective of this work is to construct and validate a longitudinal biometric dataset of neonates and infants. This dataset will comprise multiple fingerprint captures per subject, collected at various time points from birth through the first year of life. Such a resource is essential not only for understanding the aging effect in infant biometrics, but also for enabling the development and evaluation of advanced deep learning algorithms and identity management strategies tailored to the unique challenges of neonatal identification. Ultimately, this initiative supports the broader goal of enhancing healthcare delivery, nutrition monitoring, and child protection programs through reliable, long-term biometric identification.

% HIPÓTESE
The underlying hypothesis of this work is that employing artificial intelligence can provide a better approach to emulating the growth and changes in the minutiae map of newborns over time. It is believed that by developing a deep learning-based growth model that includes as input the original fingerprint, the segmented fingerprint (valley and ridge relation), and the minutiae map, it is possible to mitigate the impact of the ageing effect on biometric features, thereby providing a more accurate and adaptable representation of newborn development than simply applying scaling factors as reported by the literature. 

To achieve the proposed goal, the following steps are undertaken:

\begin{itemize}
    \item \textbf{Design of Data Collection Protocol:} A comprehensive longitudinal acquisition protocol is established, considering ethical guidelines, consent procedures, environmental factors, and appropriate intervals for data collection throughout the first year of life.

    \item \textbf{Use of specific Biometric Capture System:} A specialized fingerprint acquisition system is used to accommodate the anatomical and physiological characteristics of neonates and infants, ensuring minimal discomfort and high-quality image capture.

    \item \textbf{Data Acquisition:} Biometric data is collected from a cohort of neonates at multiple time points, beginning shortly after birth and extending to 12 months of age. Each subject contributes multiple fingerprint samples per session to assess intra-session and inter-session variability.

    \item \textbf{Data Annotation and Curation:} Collected samples are annotated with metadata including age at capture, gestational age, sex, and relevant health information, followed by rigorous quality control and curation.

\end{itemize}

This structured approach lays the foundation for scalable and reliable biometric identification systems for infants, with significant implications for healthcare, nutrition, and child protection efforts in low-resource settings. This is currently being conducted in the scope of a research group\footnote{Information Processing Research Group - Federal University of Technology - Paraná (in Portuguese: \textit{ Grupo de Pesquisas em Processamento de Informação na Universidade Tecnológica Federal do Paraná - UTFPR}). The research has been approved with a Certificate of Presentation of Ethical Appreciation 73791023.7.0000.0177 at the Brazil Platform.}.

\section{Related works}\label{sec:related}

The retrieved studies indicate several neonatal fingerprint datasets that were available in the literature.  Table \ref{tbl:multidataset} shows the longitudinal datasets, i.e., the datasets obtained with follow-up collection sessions, with more than one collection of fingerprints for the same finger across time. Age and time-lapse information are expressed in the same time unit that the original studies\footnote{Ages/time lapses are expressed in intervals using the same time unit presented in the original study: years (y), months (m), weeks (w), days (d), and hours (h). For some studies, the average is available and shown in parenthesis.}. 
Each dataset can have several age intervals with different numbers of people and total images. When available, we also show the number of images per person, the description of the collected biometric feature\footnote{For clarity purposes, we considered left (L), right (R), finger (F), thumb (T), index finger (If), middle finger (Mf), ring finger (Rf), little finger (Lf), session (ss), template (tp), and scanner (sc).}, the number of collected impressions, and where the collection was performed.

\begin{table}[h]
\begin{tblr}{
width=\textwidth,
colspec = {X[c,m]X[c,m]X[c,m]X[c,m]X[c,m]X[c,m]},
row{1-Z} = {font=\scriptsize},
abovesep = -2pt,
belowsep = -2pt,
row{2,6,7,8} = {abovesep = 0pt},
}
\hline[.8pt] 
\textbf{Study} & \textbf{Sessions} & \textbf{Age} & \textbf{Time lapse} & \textbf{People per ss} & \textbf{Total images}\\\hline
\cite{galton1899finger} & 6 & 9d-1m & {0-1m, 3.5-6m, 10-12m, 1m, 3y} & 1 & $>$=60\\
\cite{schneider2010quantifying} & 3 & 1-18y & {1y, 1y} & {308, 186, 123} & {1232, 744, 492}\\
\cite{guenter2013fingerprint} & 2 & 0-11y & some y & 1632 & 6528 \\
\cite{jain2014recognizing} & 5 & 0-4y & 1w & 20 & 1600 \\
\cite{jain2016fingerprint} & 4 & 0-5y & {6m, 4m, 2m} & {204, 167, 180, 178} & {1224, 2004, 2160, 2136}\\
\cite{jain2016fingerprint} & 3 & 0-42 (6.1)w & {4m, 2m} & {65, 52, 50} & {780, 624, 600} \\
\cite{jain2016fingerprint} & 2 & 0-42 (7.6)w & 2m & {40, 30} & {480, 360} \\
{\cite{jain2016giving} \cite{koda2016advances}} & 2 & 0-6m & 2-4d & 66  & 792\\
\cite{camacho2017recognizing} & 2 & 0-10y & 4.6-20m & $>$= 45395  & 45395 pairs\\
\cite{macharia2017feasibility} & 3 & {0-3m, 4-6m, 7-9m, 10-12m} & {1m, 1m} & {20, 20, 20, 20} & {176, 130, 153, 58}\\
{\cite{basak2017multimodal} \cite{basak2017biometric} \cite{ma2017telebiometric}} & 2 & 18m-4y & $>$=4m & {119, 108} & {5950, 5400}\\
\cite{haraksim2019fingerprint} & 2 & {5-17y, 18-25y} & 0-6y & {139566, 73056}  \\
\cite{galbally2018fingerprint} & 2 & {0-17y, 18-25y, 65-98y} & 0-8y & 265341 & {222060, 147985, 56291} \\
{\cite{beslay2018automatic}\cite{galbally2018study}} & 2 & {0-17y, 18-25y, 65-98y} & 0-8y & {102600, 36528, 16959} & 312174\\
\cite{saggese2019biometric} & 2 & $<$6m & 0-388 (38)d & 504 & 44953\\
\cite{j2019infant} & 3 & 0-3m & {2-3d, 3m} & {76 (1ss),  40 (2ss)\\ 78 (3ss)} & 1560 \\
\cite{koda2019development} & Several & 2h-24h &  & 50 & $>$3000 \\
\cite{preciozzi2020fingerprint} & 2 & 5y & 10-13y & 10000 &59004 \\
\cite{nugroho2021image} & 7 & Newly born & 1m & 1 & 280 \\
\cite{engelsma2021infant} & 4 & 0-3m & 3-12m & {127 (1ss), 109 (2ss), 54 (3ss), 25 (4ss)}  & 3071 \\
\cite{ajudesign} & 2 & 0-6m & 1w & {15, 22} & {240, 352} \\\hline
\end{tblr}
\caption{Multi-session collection datasets.}
\label{tbl:multidataset}
\end{table}

Regarding age, studies have split children for image collection into different intervals with no apparent pattern. While some studies have focused on newly-born children \cite{koda2022fundamental,nugroho2021image}, others have collected images along the first months of life \cite{galton1899finger,jain2016giving,koda2016advances,saggese2019biometric,engelsma2021infant, jain2014recognizing,engelsma2018fingerprint}, or the first years, or even adulthood \cite{joun2003experimental, schneider2010quantifying, jain2016fingerprint, camacho2017recognizing, haraksim2019fingerprint,galbally2018fingerprint,haraksim2019fingerprint}. The time lapse between collections in the longitudinal studies has also varied from days \cite{jain2016giving,koda2016advances,j2019infant} to months and years \cite{schneider2010quantifying,haraksim2019fingerprint,galbally2018fingerprint,preciozzi2020fingerprint}. The results show that the studies regarding younger children have reached at most a few hundred people \cite{mukoya2021feasibility,schneider2010quantifying,jain2016fingerprint} while the ones including older children have achieved the order of thousands of images \cite{beslay2018automatic,galbally2018study,preciozzi2020fingerprint}, revealing the difficulties inherent to the process of collecting neonatal children's fingerprints. The number of impressions taken ranges from one \cite{uhl2009comparing, engelsma2018fingerprint, koda2022fundamental,schneider2010quantifying,guenter2013fingerprint,galbally2018study} to ten \cite{nugroho2021image}. The fingers used in the collection are normally the thumb and index fingers, but some studies have used all ten fingers \cite{dharma2018level,basak2017biometric,basak2017multimodal}.

We remark that most multi-session studies (for recollection of fingerprints from the same individual) have considered only two sessions. The maximum number of sessions found in the literature was seven in \cite{nugroho2021image}, but only one child was accompanied in this case. The low number of sessions suggests the difficulty in tracking the same person across the years, especially in the neonate scenario. These arguments confirm a significant gap in the literature since some studies, especially those focusing on emulating the growth of the fingerprint across the years, would only be possible upon databases that include many recollections from the same individuals at different points in time.

\section{Dataset collection protocol}\label{sec:method}

\begin{figure}
    \centering
    \includegraphics[width=0.5\linewidth]{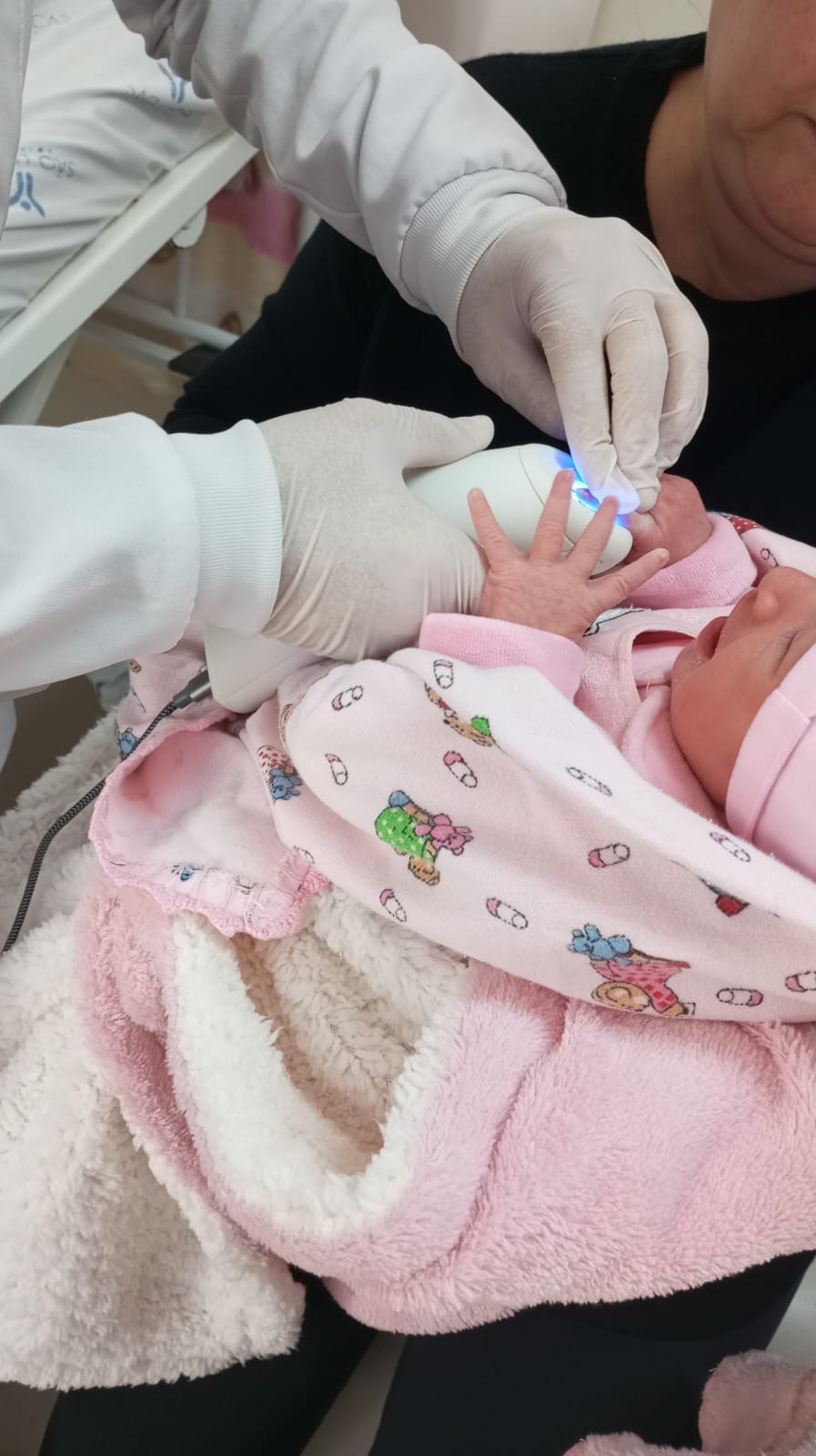}
    \caption{Collection process}
    \label{fig:collect}
\end{figure}

 The study is conducted by the 
Information Processing Research Group - Neonatal Biometrics from the Federal University of Technology – Parana, in Brazil\footnote{https://sites.google.com/view/utfprbiometria}.
The data collection process was carefully designed to ensure both scientific rigor and ethical compliance with Brazilian regulations.
It begins with trained field researchers approaching mothers who have recently given birth at partnering hospitals. During this initial interaction, the project is introduced, including its objectives, the research group's mission, and its collaboration with academic institutions. Mothers who agree to participate are asked to sign the Informed Consent Form, as required by Brazilian law through the Plataforma Brasil ethics framework. 
Once consent is obtained, biometric data collection is initiated.Figure \ref{fig:collect} shows the collection process.

The process involves capturing the fingerprints of all ten fingers of the newborn, starting from the left pinky finger to the right pinky finger. The scans are performed using a high-resolution (5000 ppi) fingerprint scanner developed by Infant.ID \cite{infantid_2025_infantid}, a pioneer company specializing in infant biometrics. Through strategic partnerships with institutions, the company has implemented biometric solutions that prevent newborn misidentifications and enable the civil registration of children aged 0 to 5 years. Its technology has proven critical in reducing sub-registration rates, supporting social protection initiatives, and enhancing security in both hospital and governmental contexts. This scanner is specifically designed to accommodate the delicate and small skin surfaces of neonates. Its imaging system uses a high-density optical sensor capable of capturing the fine ridge details present even in newborn fingerprints.

In addition to the full ten-finger capture, the pinch fingers—thumbs and index fingers—are recaptured to allow for further analysis of variability and to ensure redundancy. Initially, these captures were taken as still images. However, after the first few months of the project, the protocol was improved to use short video recordings. This change enables the extraction of multiple frames per capture session, thereby increasing the chances of obtaining high-quality images despite common challenges such as infant movement.

All collected data is securely stored in a central repository, along with corresponding metadata, including the time and conditions of acquisition. A report about the behavior of the baby during the collection, and conditions such as luminosity, temperature, and humidity, is filled out by the field researchers. They also gather information about the gestational period of the mother, such as medications, procedures, comorbidities, and other health information.

Following the initial hospital collection, the research team maintains contact with the participating families. Phone calls are made to schedule follow-up visits at the families' homes, as part of the longitudinal data acquisition protocol. These follow-ups are scheduled at key developmental milestones: 7 days, 14 days, 1 month, 2 months, 3 months, 6 months, and 12 months of age.

During each follow-up visit, the same procedure is carried out: all ten fingers are scanned, and the pinch fingers are recaptured. The field researchers travel to the homes, bringing portable biometric equipment. Prior to each session, the scanner and the infant’s fingers are carefully sanitized to ensure hygiene and safety, both in the hospital and during home visits.

This rigorous and ethically sound process ensures high-quality, longitudinal biometric data, which is essential for advancing research in infant biometric identification, fingerprint permanence, and the development of robust machine learning models tailored to this unique population.
\section{Results}\label{sec:results}

Since the inception of the project in December 2023 through April 2025, more than 450 children have been enrolled. As part of the longitudinal biometric data collection effort, a total of 829 home-based re-collection sessions have been successfully conducted. These sessions are distributed across key developmental milestones, as summarized in Table~\ref{tab:recollection_summary}.

Each biometric session includes 14 individual fingerprint videos per child—10 fingers plus repeated captures of thumbs and index fingers. Given that each video file is approximately 80 MB in size, the project has accumulated a total of 9,940 biometric video files, amounting to roughly 776.56 GB of biometric data.

\begin{table}[h!]
\centering
\caption{Summary of Biometric Collection Sessions}
\label{tab:recollection_summary}
\begin{tabular}{lc}
\toprule
\textbf{Age at collection} & \textbf{Number of Sessions} \\
\midrule
0 days   & 452 \\ 
7 days   & 164 \\
14 days  & 147 \\
1 month  & 140 \\
1 month  & 118 \\
3 months & 113 \\
6 months & 101 \\
1 year   & 49 \\
\midrule
\textbf{Total} & \textbf{1,284} \\
\bottomrule
\end{tabular}
\end{table}

This growing repository is one of the most comprehensive longitudinal neonatal biometric datasets to date and is crucial for advancing studies in infant fingerprint permanence and machine learning models for reliable early-life identification.

Below, we present a series of images of one child; each column represents one child’s fingerprint at different ages from birth through 6 months. Within each column:
The top row shows the raw fingerprint image captured at that age.
The middle row shows the ridge pattern segmentation, highlighting the flow of ridges and valleys.
The bottom row displays the minutiae points (e.g., ridge endings and bifurcations) detected in the fingerprint.

All images are scaled to the same ratio, so the relative size differences reflect actual growth of the finger. Visually, we can see that as the child grows, the fingerprint enlarges, but the overall pattern of ridges and valleys remains consistent – the fingerprints do not change in their fundamental pattern, only in scale. For example, a tiny loop or whorl present in the newborn’s fingerprint is still present in the 6-month-old fingerprint, just larger. The minutiae (small distinguishing features) observed at later ages correspond to those present at birth, merely spaced farther apart due to the growth of the finger.

\begin{figure}
    \centering
    \includegraphics[width=1\linewidth]{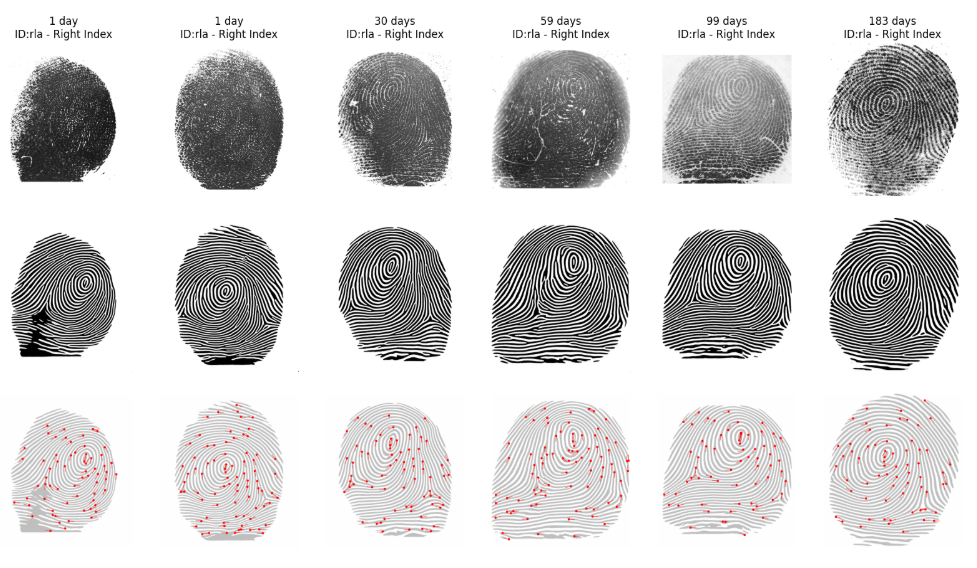}
    \caption{Example of prints of one child across time, with the respective segmentation and minutiae map}
    \label{fig:dedos}
\end{figure}

\section{Conclusion}
This study addresses a critical gap in the field of neonatal biometrics by proposing the development of a longitudinal fingerprint dataset for infants, collected through a rigorous and ethically grounded protocol. The biometric data, gathered at multiple developmental milestones from over 450 neonates, provides a robust foundation for analyzing the complex interplay between physiological growth and fingerprint permanence. By incorporating high-resolution imaging, repeated captures, and detailed metadata, this dataset enables a deeper understanding of the aging effect in infant biometrics—an issue that current literature has largely overlooked due to the lack of long-term data.

Moreover, this work introduces a novel research direction: the application of deep learning to emulate growth-related distortions in the minutiae map over time. By leveraging not only the minutiae but also the ridge-valley structures, this approach promises to move beyond the oversimplified linear scaling models found in previous studies. The ultimate goal is to support the creation of accurate and resilient biometric identification systems tailored specifically to neonates and infants.

Such systems hold the potential to significantly enhance global child health, nutrition monitoring, and protection initiatives, particularly in low-resource settings. By enabling timely vaccination, reliable tracking of growth metrics, and secure identification from birth, this research contributes to the broader vision of improving early childhood care and safeguarding vulnerable populations through technological innovation.
\section{Acknowledgements}

We extend our sincere gratitude to InfantID for their invaluable operational support to our neonatal biometrics research group at UTFPR. Their partnership has been instrumental in advancing our exploration of biometric technologies in neonatal care. This study was financed in part by the Coordenação de Aperfeiçoamento de Pessoal de Nível Superior - Brasil (CAPES) - Finance Code 001, and by the Conselho Nacional de Desenvolvimento Científico e Tecnológico (CNPQ).

\bibliography{main}

\end{document}